\documentclass[10pt,twocolumn,letterpaper]{article}

\usepackage{cvpr}              

\usepackage[dvipsnames]{xcolor}

\definecolor{cvprblue}{rgb}{0.21,0.49,0.74}
\usepackage[pagebackref]{hyperref}
\hypersetup{colorlinks,
            breaklinks,
            linkcolor=cvprblue,
            citecolor=cvprblue,
            urlcolor=cvprblue}
        
\newcommand{\cref}[2]{\hyperref[#2]{#1~\ref*{#2}}}
\newcommand{\colref}[2]{\hyperref[#2]{#1~\ref*{#2}}}

\newcommand{\figref}[1]{\colref{Figure}{#1}}

\newcommand{\secref}[1]{\colref{Section}{#1}}
\newcommand{\tabref}[1]{\colref{Table}{#1}}
\newcommand{\coloredref}[2]{\hyperref[#2]{#1~\ref*{#2}}}
\newcommand{\coloredsubref}[3]{\hyperref[#2]{#1~\ref*{#2}{#3}}}
\newcommand{\Algref}[1]{\hyperref[#1]{Algorithm~\ref*{#1}}}

\def\paperID{*****} 
\def\confName{CVPR}
\def\confYear{2024}

\title{3D Reconstruction of Protein Structures from Multi-view AFM Images using Neural Radiance Fields (NeRFs)}

\author{Jaydeep Rade
\and Ethan Herron
\and Soumik Sarkar
\and Anwesha Sarkar
\and Adarsh Krishnamurthy
\\
Iowa State University\\
Ames, Iowa, USA\\
{\tt\small (jrrade, edherron, soumiks, anweshas, adarsh)@iastate.edu}
}

\begin{document}
\maketitle
\begin{abstract}
Recent advancements in deep learning for predicting 3D protein structures have shown promise, particularly when leveraging inputs like protein sequences and Cryo-Electron microscopy (Cryo-EM) images. However, these techniques often fall short when predicting the structures of protein complexes (PCs), which involve multiple proteins. In our study, we investigate using atomic force microscopy (AFM) combined with deep learning to predict the 3D structures of PCs. AFM generates height maps that depict the PCs in various random orientations, providing a rich information for training a neural network to predict the 3D structures. We then employ the pre-trained UpFusion model (which utilizes a conditional diffusion model for synthesizing novel views) to train an instance-specific NeRF model for 3D reconstruction. The performance of UpFusion is evaluated through zero-shot predictions of 3D protein structures using AFM images. The challenge, however, lies in the time-intensive and impractical nature of collecting actual AFM images. To address this, we use a virtual AFM imaging process that transforms a `PDB' protein file into multi-view 2D virtual AFM images via volume rendering techniques. We extensively validate the UpFusion architecture using both virtual and actual multi-view AFM images. Our results include a comparison of structures predicted with varying numbers of views and different sets of views. This novel approach holds significant potential for enhancing the accuracy of protein complex structure predictions with further fine-tuning of the UpFusion network.
\end{abstract}

\section{Introduction}
\label{sec:intro}

\begin{figure}[b!]
  \centering
  \includegraphics[width=0.99\linewidth, trim={4in 3.5in 4in 3.0in},clip]{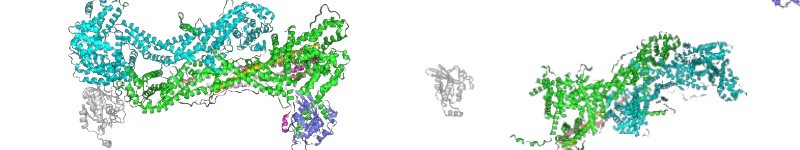}
   \caption{AlphaFold2 multimer (right) is unable to accurately predict the structure of the WRC protein complex (left).}
   \label{fig:AlphaFold_prediction_WRAC}
\end{figure}

\begin{figure*}[t!]
  \centering
  \includegraphics[width=0.99\linewidth, trim={0.5in 1in 0.5in 1in},clip]{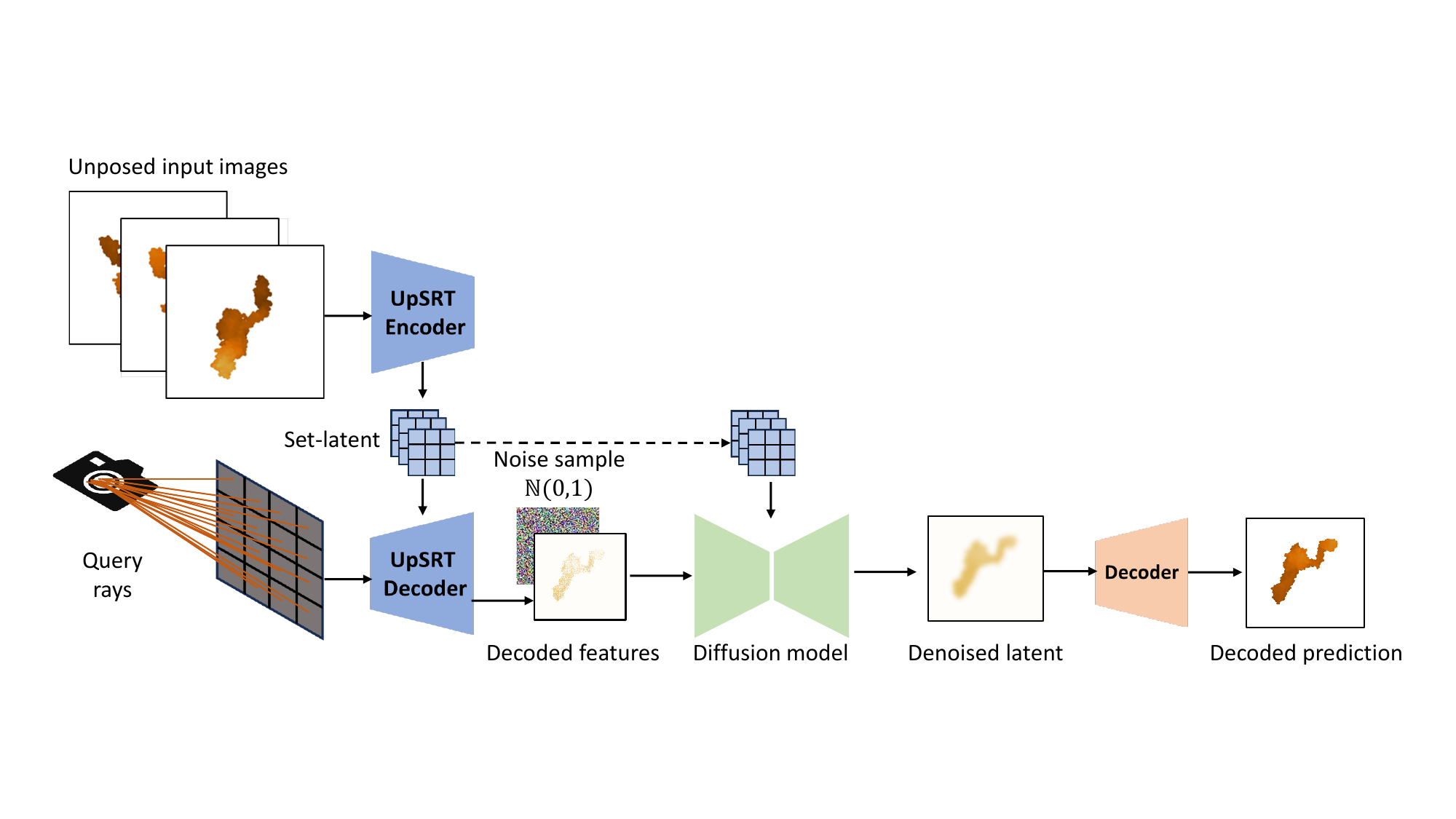}
   \caption{UpFusion architecture for synthesizing novel views from unposed input images.}
   \label{fig:architecture}
\end{figure*}

Accurately predicting the 3D structure of proteins from their amino acid sequences has been a formidable challenge for the last 50 years~\citep{PF2008Dill}. Predicting the 3D structure of a protein is critically important because the structure dictates its function~\citep{PF20212Dill}. Over the years, researchers have made use of and refined experimental techniques such as X-ray crystallography, nuclear magnetic resonance (NMR) spectroscopy, and cryo-electron microscopy (cryo-EM) to ascertain the 3D structures of proteins~\citep{NMR2001Kurt, Thompson2020AdvancesIM,cryoEM2015Bai, Jasklski2014ABH}. These methodological advancements have significantly increased the structural data available for numerous proteins. Concurrently, developments in deep learning (DL) have significantly influenced the field of protein structure prediction, exemplified by the creation of AlphaFold2~\citep{AF22021Jumper}, ESMFold~\citep{esm2fold} and RoseTTAFold~\citep{RF2021Baek}. The integration of DL with breakthroughs in cryo-EM imaging, particularly in single-particle analysis, has facilitated the elucidation of protein structures and their complexes. Nonetheless, cryo-EM images often suffer from high noise levels, with signal-to-noise ratios reaching as low as -20 dB~\citep{Gupta2020MultiCryoGANRO}. Furthermore, conducting cryo-EM experiments is costly, with expenses potentially reaching up to \$1,500 for each protein structure analyzed~\citep{Cianfroco2015}.

An alternative approach to deducing protein structures involves establishing a relationship between a protein's amino acid sequence and its corresponding structure. This strategy encompasses techniques like Position Specific Scoring Matrices (PSSM)~\citep{wang2021PSSP}, which calculate the frequency distribution of mutations at each residue position, and Multiple Sequence Alignment (MSA), which identifies co-evolutionary patterns among sequences~\citep{Senior2020ImprovedPS, Yang2020}. Notably, AlphaFold2 leverages comprehensive MSAs and a template-based prediction process. However, its effectiveness diminishes when it comes to predicting the structures of protein complexes, where multiple proteins interact closely. For instance, despite being tailored for the prediction of protein complexes, AlphaFold2-multimer~\citep{AF2Multi2022Evans} struggles to accurately forecast the structure of the WRC protein complex~\citep{koronakis2011wave,rottner2021wave,ismail2009wave,chen2014wave} as seen in \figref{fig:AlphaFold_prediction_WRAC}. Given the limitations of both \textit{in silico} methods, which can fail to predict the structure of protein complexes in certain instances, and the high costs and lengthy processes associated with Cryo-EM, we suggest utilizing Deep Learning (DL) models trained on images obtained through Atomic Force Microscopy (AFM). AFM~\citep{sarkar2019live,sarkar2015interaction,sarkar2022biosensing,sarkar2022multimodal,stanley2023effects,jones2017revisiting}, falls under the umbrella of scanning probe microscopy and offers a non-destructive means to achieve high-resolution imaging of proteins under physiological conditions. This technique avoids the need for extensive sample preparation (such as freezing, drying, or dye-tagging), which can compromise the integrity of these delicate samples.

\begin{figure*}[t!]
  \centering
  \includegraphics[width=0.65\linewidth, trim={3.0in 2.75in 3.2in 2.5in},clip]{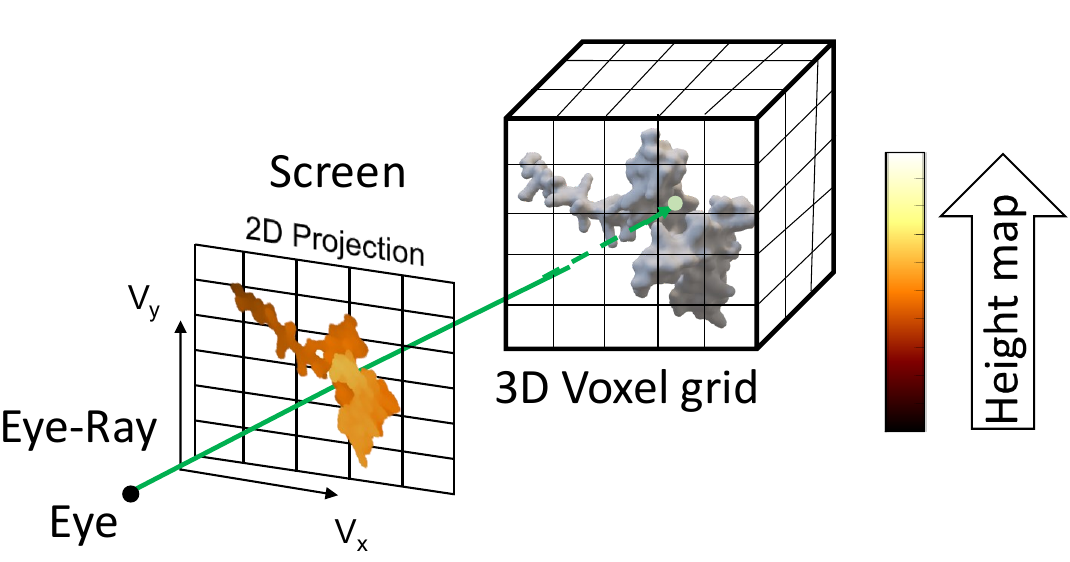}
   \caption{Virtual AFM: GPU-accelerated volume rendering of 3D voxelized protein structure to generate AFM-like virtual 2D images.}
   \label{fig:volumeRendering}
\end{figure*}

With these advantages of AFM, we propose a novel approach to infer the 3D structure of protein complexes using a neural network trained using multi-view AFM images. We use UpFusion~\citep{kani2023upfusion} neural network architecture, which has demonstrated superior performance in synthesizing novel views and inferring 3D structures. Specifically, it can exploit its ability to reconstruct 3D models from unposed images and generalize beyond the training categories. We explain the details of the architecture in \figref{sec:method}. We assess the UpFusion architecture through a zero-shot approach for predicting protein structures using AFM images. Despite having multiple actual AFM images of a WRC protein complex, using just one protein for validation does not sufficiently verify this method. To validate this method extensively we take use of a virtual AFM imaging pipeline. Even though AFM is not as expensive as Cryo-EM, experimentally acquiring the AFM images for large number proteins is not practical. Hence, using a volume rendering technique, a virtual AFM imaging pipeline to generate multiple views of the proteins. We discuss this virtual imaging process in detail in \secref{sec:virtualAFM}. We show our preliminary results in \secref{sec:results}. Finally, we conclude in \secref{sec:conclusion} with some future directions.

\section{Method}
\label{sec:method}

We leverage the pre-trained neural network called UpFusion, a novel system designed to perform two main tasks: novel view synthesis and inferring 3D representations of objects from a sparse set of reference images without corresponding pose information. This is a significant departure from traditional sparse-view 3D inference methods that rely on geometrically aggregating information from input views using camera poses~\citep{Yao2018MVSNet, yu2021pixelnerf,Chan2023GenerativeNV,zhou2023sparsefusion}, which are often unavailable or inaccurate in real-world scenarios. UpFusion overcomes this challenge by employing a conditional generative model that leverages the available images as context without requiring explicit camera poses to synthesize novel views.

The core methodology of UpFusion integrates two key components: a scene-level transformer and a denoising diffusion model. The transformer, specifically an Unposed Scene Representation Transformer (UpSRT)~\citep{Mehdi2022UpSRT}, is utilized to infer query-view aligned features by implicitly incorporating all available input images as context. Conditioning a diffusion model with the internal representations of UpSRT can generate novel views from sparse data, thus enabling probabilistic sparse view synthesis~\citep{Rombach2022, zhang2023controlnet}. To enhance the specificity and relevance of the generated views, UpFusion incorporates `shortcuts' through attention mechanisms in the diffusion process, allowing direct access to input view features during generation. 

Although the conditional diffusion model we propose can produce high-quality renderings from query views, it fails to ensure 3D consistency across the generated views. To achieve a coherent 3D representation (Neural radiance field or NeRF) from the distribution inferred over new views, it refines this process by optimizing an instance-specific neural representation~\citep{mueller2022instant}. This optimization draws inspiration from SparseFusion~\citep{zhou2023sparsefusion}, which focuses on identifying neural 3D structures by enhancing the likelihood of their renderings. This is achieved by adapting the Score Distillation Sampling (SDS)~\citep{poole2023dreamfusion} loss for use with view-conditioned generative models.

The training of UpFusion involved a multi-stage procedure, starting with the training of the UpSRT model using a reconstruction loss on the colors predicted for queried rays based on a set of reference images, followed by training the denoising diffusion model with conditioning information from the pre-trained UpSRT. After training the diffusion model, it derives a 3D representation of an object through the optimization of an Instant-NGP~\citep{mueller2022instant, torch-ngp}. The NeRF is trained for 3000 iterations, which takes a around an hour on an A100 GPU for a single protein sample. This process highlights the importance of a systematic approach to training complex models that combine the strengths of transformers and diffusion models for 3D inference and novel view synthesis.

\section{Virtual AFM}
\label{sec:virtualAFM}

To conduct extensive evaluation of the UpFusion architecture we need AFM images for sufficiently large number of proteins. Acquiring actual AFM images for a large number of proteins is an impractical task. To solve this, we take use of a `virtual AFM' workflow for generating virtual multi-view AFM images from protein files in `PDB' format. We utilize the `AlphaFold DB'~\citep{Varadi2021AFDB}, a database comprised of predictions made by the AlphaFold2~\citep{AF22021Jumper} model on Swiss-Prot entries. Despite facing challenges in predicting protein complexes, AlphaFold2 excels in predicting the structure of individual proteins. Subsequently, we employ PyMol, a software frequently utilized for protein visualization, to convert the `PDB' format structures into three-dimensional mesh files in `OBJ' format. It is possible to visualize proteins in various visual forms, including surfaces, spheres, sticks, cartoons, or ribbons. For our purposes, we choose to export the `OBJ' files depicting the surface representations of the proteins. Following this, we apply a GPU-accelerated algorithm~\citep{young2018voxel} to voxelize the 3D mesh. We set the voxelization resolution at 256 voxels along the protein's longest dimension and apply zero padding to the other dimensions to ensure the voxelized structure achieves a uniform resolution of \(256 \times 256 \times 256\). 

AFM generates a topographic image by scanning the protein sample's surface, delivering precise height map details. The one of the most common colormaps used in AFM images is the `hot' colormap, which ranges from black (representing the lowest points) through reds and yellows to white (representing the highest points). This colormap is often used because it effectively highlights variations in height across the surface being imaged, making it easier to discern topographical features. In our methodology, we employ a GPU-accelerated volume rendering~\citep{young2018gpvol} technique to produce 2D height-map projections from the voxelized 3D structure of proteins, mimicking the AFM imaging process as illustrated in \ref{fig:volumeRendering}. During AFM imaging of proteins, the protein molecules rest in various orientations on the AFM platform, producing diverse views of the protein. To gather a multiple random views through virtual AFM, the 3D structure undergoes random rotations within the three-dimensional space, resulting in a collection of multi-view images. For each protein structure we save 25 random views, saved in a standard image format. 
Moreover, our study extends to include high-resolution imaging at the nanometer scale of single molecules from the WRC protein complex~\citep{koronakis2011wave,rottner2021wave,ismail2009wave,chen2014wave,Chen2022WRC}. Virtual AFM images of the WRC complex are also created. 


\section{Results and Discussion}
\label{sec:results}

\begin{figure}[t!]
  \centering
  \includegraphics[width=0.99\linewidth, trim={1.3in 0.75in 1.3in 0.75in},clip]{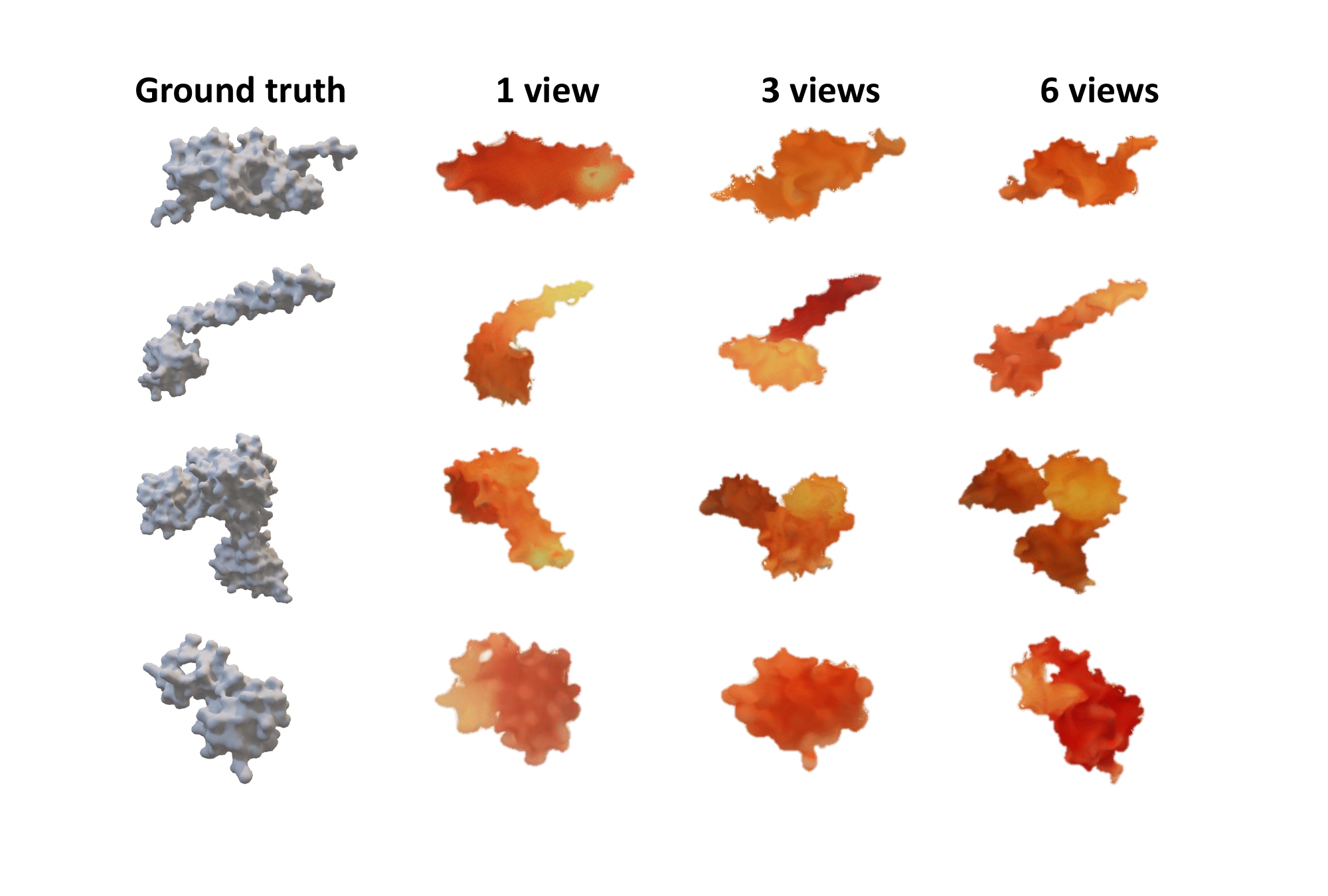}
   \caption{Comparing the predictions corresponding to different number of input views.}
   \label{fig:diff_num_views}
\end{figure}

\begin{figure}[t!]
  \centering
  \includegraphics[width=0.99\linewidth, trim={2.2in 1.25in 2.2in 1.25in},clip]{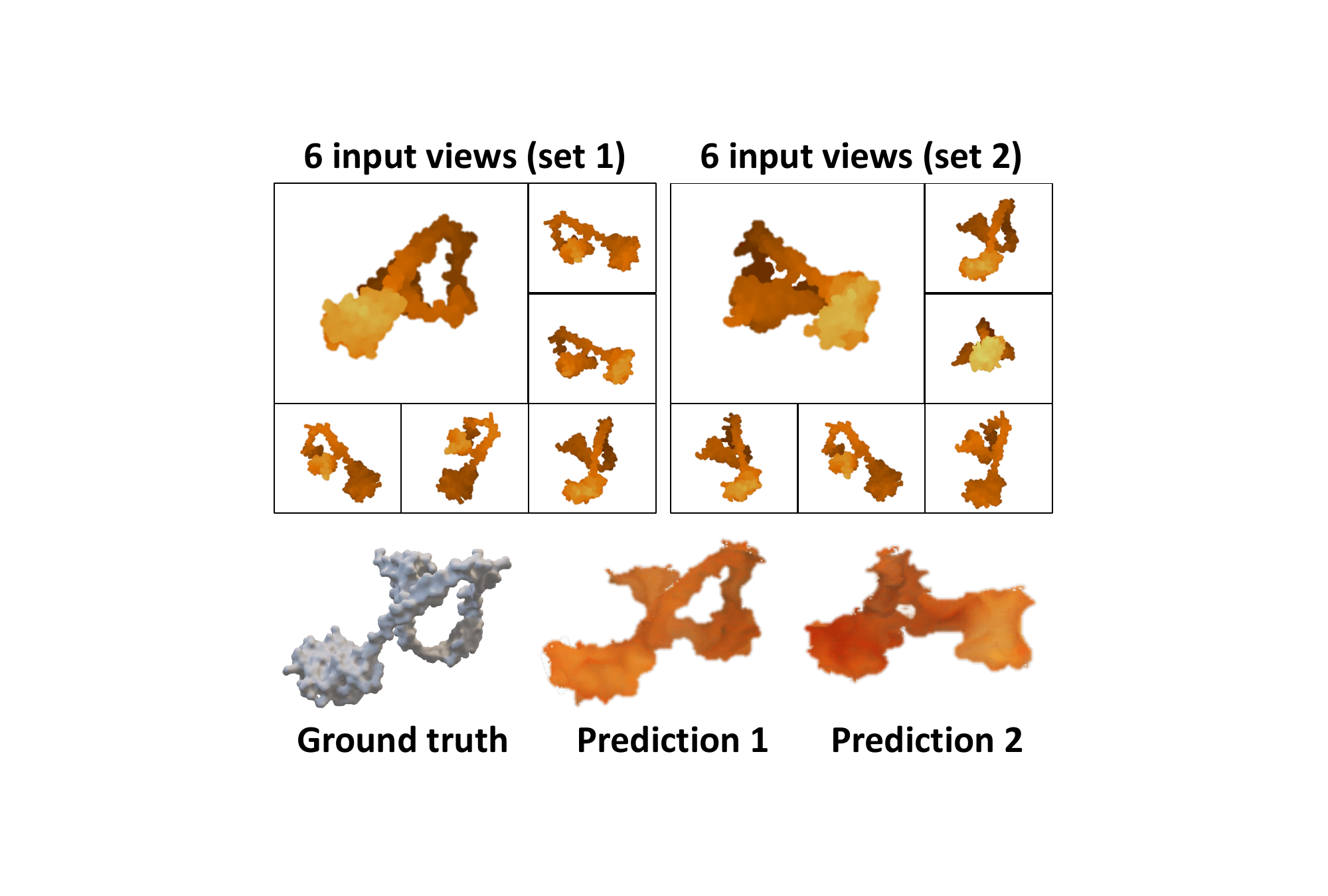}
   \caption{Comparing different set of input views for same protein structure, prediction 1 and 2 corresponds to input set 1 and 2, respectively.}
   \label{fig:diff_set_views}
\end{figure}

We evaluate the zero-shot capability of 3D reconstructions of the UpFusion network using both virtual and real unposed AFM images. In \figref{fig:diff_num_views}, we present the outcomes from utilizing varying numbers of input views (e.g., 1V, 3V, and 6V), noting an improvement in 3D reconstruction quality with an increase in the number of input images. For quantitative assessment, we employ established metrics for image reconstruction, including Peak Signal-to-Noise Ratio (PSNR), Structural Similarity Index (SSIM)~\citep{Wang2004SSIM}, and Learned Perceptual Image Patch Similarity (LPIPS)~\citep{Zhang2018LPIPS}. We perform a numerical analysis of the novel view synthesis for each protein depicted in \figref{fig:diff_num_views}. Subsequently, we calculate the average of each metric across the protein samples and present the results in \tabref{tab:metrics}. Consistent with the trends observed in visualized predictions, we find that an increased number of input views enhances the quality of the synthesized views.

We also investigate how the choice of input views impacts prediction accuracy. To do this, we generate the 3D predictions by feeding different sets of views for a single protein. In this approach, we form two sets of various views of a protein as illustrated in \figref{fig:diff_set_views}. It is observed that, in general, a more varied collection of views enhances the 3D reconstruction, reducing the ambiguity arising from views that appear similar.

\begin{figure}[t!]
  \centering
  \includegraphics[width=0.99\linewidth, trim={2.5in 1.0in 2.5in 1.0in},clip]{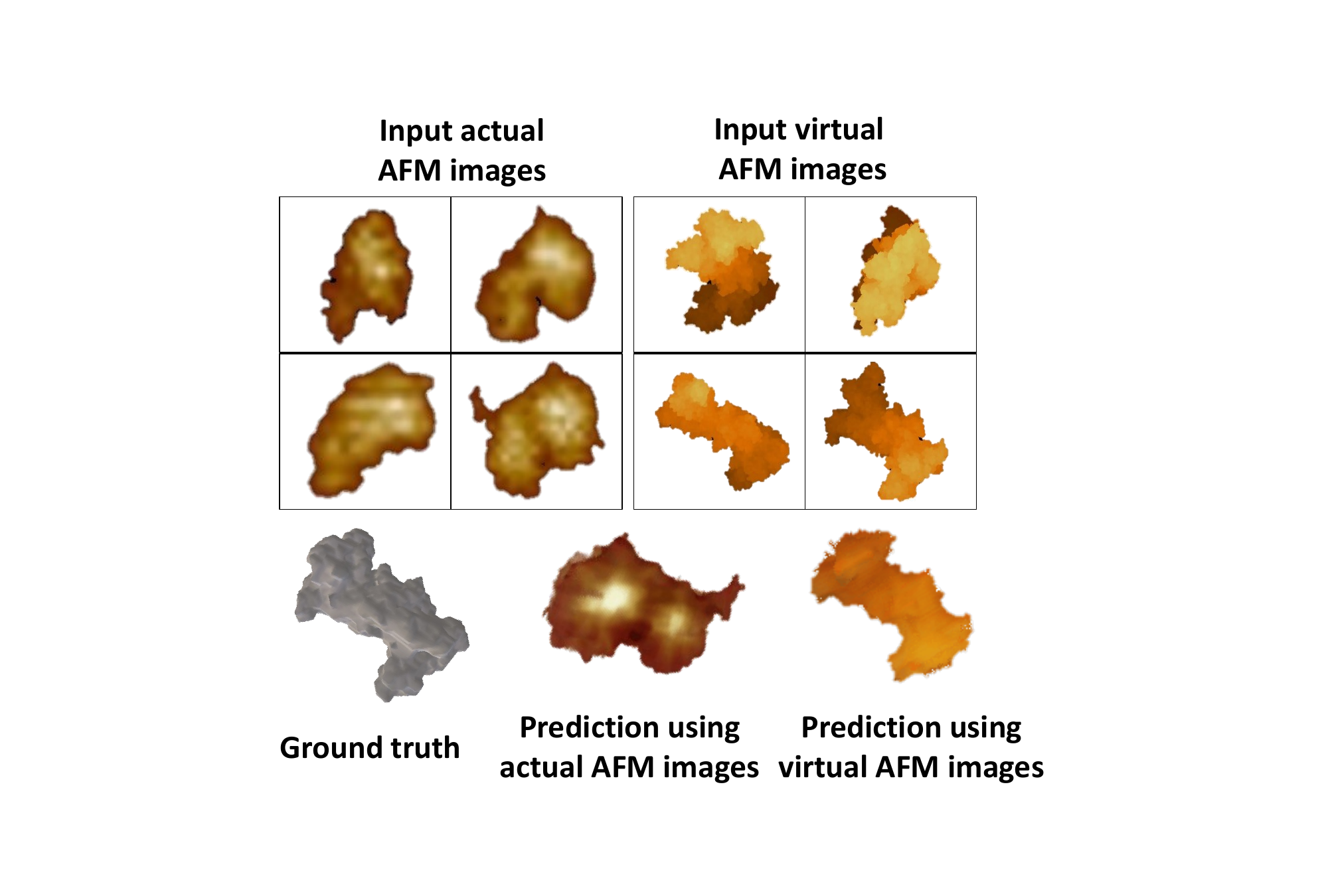}
   \caption{Prediction on WRC protein complex structure using actual and virtual AFM images.}
   \label{fig:WRAC_pred}
\end{figure}

Further, we visualize the prediction made using actual AFM images of WRC protein complex and compare it with the prediction from virtual AFM images in \figref{fig:WRAC_pred}. We see both the prediction captures the high-level shape, but the prediction from the actual AFM image fails to capture the intricate details. We anticipate this result as these are just zero-shot predictions using a network that has never been trained on AFM or protein datasets. One main discrepancy is that the color of the AFM image denotes the depth information, while the UpFusion network is trained on real-life objects with lighting that does not directly encode the depth.

\begin{table}
  \centering
  \begin{tabular}{@{}cccc@{}}
    \toprule
    $\#$ views & PSNR ($\uparrow$) & SSIM ($\uparrow$) & LPIPS ($\downarrow$)\\
    \midrule
    1 & 15.99 & 0.8512 & 0.2040 \\
    3 & 16.12 & 0.8614 & 0.1818 \\
    6 & 16.72 & 0.8737 & 0.1677 \\
    \bottomrule
  \end{tabular}
  \caption{Numerical evaluation for the novel view synthesis for different number of input views.}
  \label{tab:metrics}
\end{table}

\section{Conclusions}
\label{sec:conclusion}

In this work, we validated the deep learning framework`UpFusion', which is designed to generate the novel views and 3D NeRF reconstruction, for the task of protein structure prediction using AFM images. We utilize the virtual AFM imaging pipeline, a GPU-accelerated volume rendering technique that mimics the AFM imaging process, which facilitates the creation of virtual AFM-like images from `PDB' format protein files, which enabled us for extensive evaluation using multiple protein samples. 
We tested the efficacy of the UpFusion network by conducting zero-shot predictions on both actual and virtual AFM protein images without requiring any model fine-tuning. The UpFusion network exhibited promising capabilities in predicting the 3D structure of proteins from unposed AFM images. Future work will focus on refining the UpFusion architecture specifically for the protein structure prediction task using AFM images. We anticipate that fine-tuning with extensive dataset comprising of virtual AFM images will significantly enhance the accuracy of 3D reconstructions.


{
    \small
    \bibliographystyle{ieeenat_fullname}
    \bibliography{main}
}


\end{document}